\documentclass[letterpaper, 10 pt, journal, twoside]{IEEEtran}

\IEEEoverridecommandlockouts

\usepackage{cite}
\usepackage{ulem}
\usepackage{acronym}
\usepackage{caption}
\usepackage{leftidx}
\usepackage{listings}
\usepackage{graphicx}
\usepackage{textcomp}
\usepackage{multirow}
\usepackage{booktabs}
\usepackage{colortbl}
\usepackage{subcaption}
\usepackage{amsmath,amssymb,amsfonts}

\newcommand{\norm}[1]{\left\lVert#1\right\rVert}

\usepackage{booktabs}
\usepackage{array}
\usepackage{savesym}
\savesymbol{checkmark}
\usepackage{dingbat}
\usepackage{bbding}
\usepackage{colortbl}
\usepackage[table]{xcolor}
\usepackage{csquotes}
\usepackage{soul}
\usepackage{svg}

\usepackage{algorithm}      
\usepackage{algpseudocode}  
\usepackage{comment}
\usepackage{fontawesome5}
\usepackage{censor}
\usepackage{dblfloatfix}  

\usepackage{hyperref}
\usepackage{cleveref}

\acrodef{TP}{True Positive}
\acrodef{FP}{False Positive}
\acrodef{FN}{False Negative}
\acrodef{VSLAM}{Visual SLAM}
\acrodef{STD}{Standard Deviation}
\acrodef{ROS}{Robot Operating System}
\acrodef{RMSE}{Root Mean Square Error}
\acrodef{ATE}{Absolute Trajectory Error}
\acrodef{RANSAC}{RANdom SAmple Consensus}
\acrodef{CNN}{Convolutional Neural Network}
\acrodef{LiDAR}{Light Detection And Ranging}
\acrodef{SLAM}{Simultaneous Localization and Mapping}

\definecolor{red}{HTML}{fd8f8f}
\definecolor{greend}{HTML}{57e377}
\definecolor{greenl}{HTML}{b8fb8a}
\definecolor{lyellow}{HTML}{fefdb4}
\definecolor{orange}{HTML}{ffd5ab}

\colorlet{red}{red!50}
\colorlet{yellow}{yellow!50}
\colorlet{greenl}{greenl!50}
\colorlet{greend}{greend!50}

\newcommand{\system}{MVP-SLAM}

\title{\LARGE \bf MVP-SLAM: Multi-Camera Visual-Inertial Floorplan-Prior SLAM}

\author{
    Asier Bikandi-Noya$^{1,\dagger}$, 
    Miguel Fernandez-Cortizas$^{1,\dagger}$, 
    Muhammad Shaheer$^{1}$,\\
    Holger Voos$^{1}$, and Jose Luis Sanchez-Lopez$^{1}$%
    \thanks{$^{1}$Authors are with the Automation and Robotics Research Group, Interdisciplinary Centre for Security, Reliability, and Trust (SnT), University of Luxembourg, Luxembourg. Holger Voos is also associated with the Faculty of Science, Technology, and Medicine, University of Luxembourg, Luxembourg. 
    \tt{\small{\{asier.bikandi, miguel.fernandez, muhammad.shaheer, holger.voos, joseluis.sanchezlopez\}}@uni.lu}}
    \thanks{$^\dagger$ These authors contributed equally.}
    \thanks{*
    This work was partially funded by the Fonds National de la Recherche of Luxembourg (FNR) under the project C22/IS/17387634/DEUS and BRIDGES/2025-1/IS/19685965/BARCODE}
    \thanks{*
    For the purpose of Open Access, and in fulfillment of the obligations arising from the grant agreement, the authors have applied a Creative Commons Attribution 4.0 International (CC BY 4.0) license to any Author Accepted Manuscript version arising from this submission.}
} 

\begin{document}


\maketitle
\pagestyle{headings}

\begin{abstract}

Indoor building construction sites are demanding environments for visual SLAM, where variable lighting and repetitive, low-textured structures make the system drift over long trajectories, though structural elements such as walls remain distinguishable despite these conditions. These buildings are constructed according to their \textit{as-planned} floor plans, available from the design phase, and although the actual \textit{as-built} site can differ from this design, floor plans still provide a metric reference, both to localize the system in the building and to correct drift. Existing methods often use the floor plan to correct an already-built trajectory offline, and those that instead correct it online typically rely on depth sensors. We instead present \system{}, an online visual-inertial SLAM on two opposite-facing fisheye cameras that corrects drift from cameras alone by matching walls detected in its map to the floor plan, through a drift-aware policy. A multi-stage integration then turns each matched pair incrementally into a persistent correction, so the trajectory stays corrected and localized within the floor plan as it is built. \system{} was validated on the multi-floor construction sites of the Hilti--Trimble SLAM Challenge 2026, ranking \textbf{2nd of 22 teams} in the Localization task ($0.29$\,m mean RMSE) and \textbf{5th of 62 teams} in the SLAM task ($0.24$\,m), the top-ranked one in both tasks among those that operate online, integrate the floor plan, and localize within it.
\end{abstract}

\section{Introduction}
\label{sec:intro}

Accurate and reliable localization in indoor building construction environments is essential for automating construction workflows, such as tracking a site's progress or enabling autonomous robotic inspection~\cite{shaheer2023graph}.
Visual-inertial SLAM is a practical tool for this goal, but construction site environments pose a challenge for these systems due to variable lighting, moving workers, fast motions, and repetitive, low-textured structures~\cite{helmberger2022}, causing the estimated trajectory to accumulate drift.
Nevertheless, construction environments present structural elements, such as walls and columns, that remain distinguishable despite these conditions.

\begin{figure}[t]
    \centering
    \includegraphics[width=0.95\columnwidth]{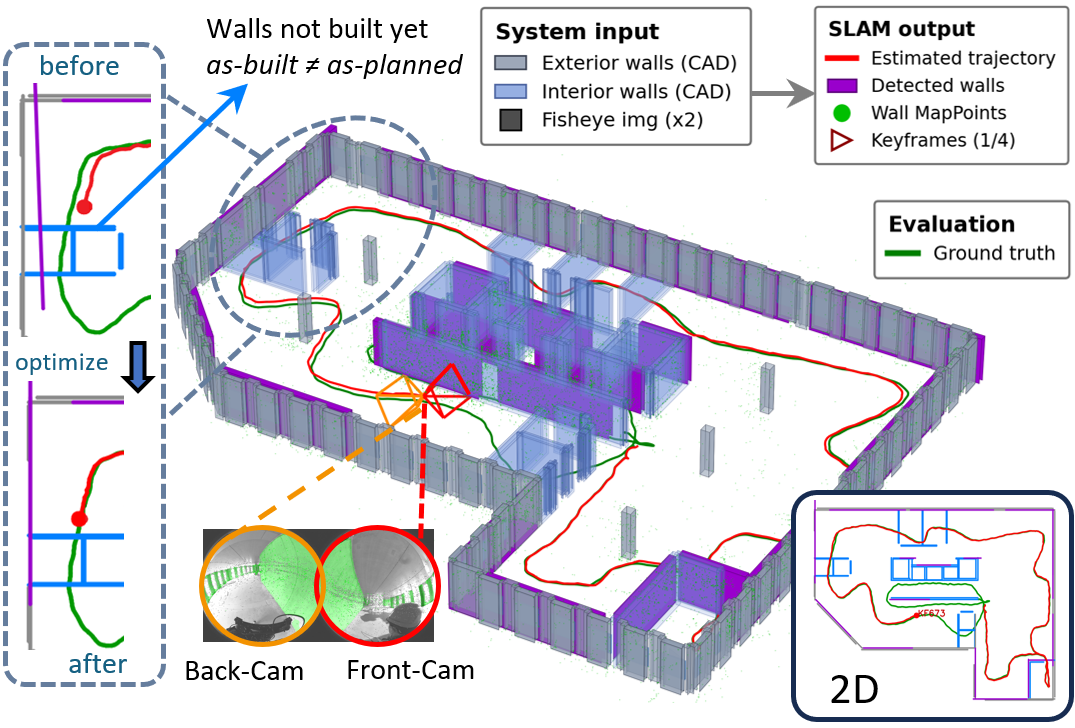}
    \caption{\system{} on a Hilti construction sequence, matching detected walls to the floor plan and correcting the estimated trajectory against ground truth.}
 \label{fig:front_image}
\end{figure}

Construction sites typically have floor plans available~\cite{mendez2018sedar}, 2D representations of these structural elements produced during the design phase that can serve as a reference to correct this drift.
In more advanced cases, these are Building Information Models (BIMs), encoding this information in more detail, though they are not universally available and are costly to process~\cite{slamchallenge2026}.
Although discrepancies exist between the \textit{as-built} site and this \textit{as-planned} floor plan~\cite{torres2023bim}, they are also used to localize in the building.
Visual localization algorithms, like Z-FLoc~\cite{umemura2026z}, localize within such a floor plan and correct drift against it, but only after a trajectory has been generated, in a post-processing step.
Onsite operation, however, needs a camera position estimate while the site is still being traversed, not only once it is complete, so the correction must happen online, as the trajectory is built, rather than recovered afterwards.

Methods offering an online camera position estimation in construction sites integrate the floor plan directly into the SLAM back-end.
These methods rely on semantic entities such as walls to establish correspondences between the \textit{as-built} site and \textit{as-planned} floor plan, incorporating each match into the back-end's graph optimization~\cite{shaheer2023graph}.
However, these systems rely on active depth or range sensors, such as anchoring LiDAR point clouds to BIMs~\cite{torres2023bim} or matching structural walls from RGB-D cameras~\cite{bikandi2025bim}, leaving visual-only settings largely unexplored.
A particular challenge for these systems is to find reliable correspondences between detected structural elements, such as walls on the \textit{as-built} site and their counterparts on the floor plan, since accumulated drift makes this association increasingly difficult as the environment grows.

The Hilti--Trimble SLAM Challenge 2026~\cite{slamchallenge2026} benchmarks such real-world construction sites and their challenging conditions, recorded with a hand-held device carrying two opposite-facing (back-to-back) fisheye cameras and an IMU, without depth sensor; the cameras cover 360$^\circ$, sharing too little overlap to act as a stereo pair.
Floor plans are also available for these sites, and one of the challenge's two tasks, \textit{Localization}, recovers the camera trajectory within the floor-plan frame from an initial pose, while its \textit{SLAM} task instead recovers it in an arbitrary frame.
Both tasks are scored on trajectory accuracy alone, so neither requires online or real-time operation.

\begin{figure*}[t]
    \centering
    \includegraphics[width=0.95\textwidth]{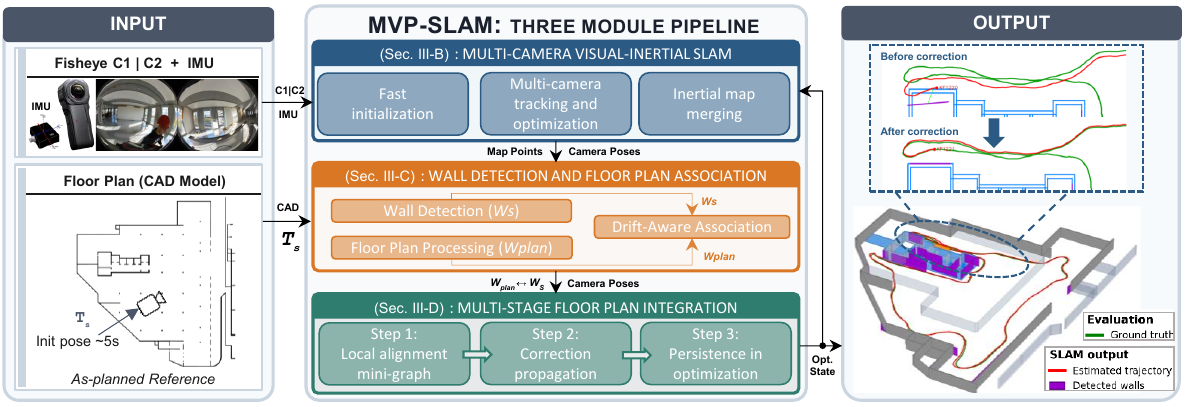}
    \caption{System architecture of \system{}. Sensor and floor-plan inputs (left) feed a three-module pipeline whose optimized state is fed back into the SLAM system as persistent plan priors, producing the plan-aligned trajectory estimate (right).
    } \label{fig:system_architecture}
\end{figure*}

To recover the trajectory in the plan's frame while correcting drift, we present \system{} (Multi-Camera Visual-Inertial Floorplan-Prior SLAM), an online visual-inertial SLAM system that leverages the building's floor plan to constrain its estimation, responding to the Hilti--Trimble SLAM Challenge 2026. Our contributions are: (i)~an online multi-camera visual-inertial SLAM system that integrates the building's floor plan to correct drift as the trajectory is built; (ii)~a semantic wall detection and drift-aware association algorithm that establishes correspondences between the \textit{as-built} map and the floor plan from cameras alone; and (iii)~a floor-plan integration method that, through a multi-stage strategy, turns each match incrementally into a bounded, persistent correction, so the map and trajectory are jointly optimized online under visual, inertial, and plan constraints.
\system{} ranked \textbf{2nd} in the Localization task of the challenge and \textbf{5th} in the SLAM task, the best-placed system in both among those that also operate online, integrate the floor plan, and localize within it.

\section{Related Work}
\label{sec:sota}

\subsubsection*{\textbf{Multi-camera visual-inertial SLAM}}
Monocular visual-inertial SLAM suits hand-held capture, but its limited field of view degrades tracking during poor lighting or occlusions.
Multi-camera setups widen visual coverage to maintain feature continuity, but whereas \textit{MAVIS}~\cite{wang2024mavis} relies on overlapping stereo pairs, purely visual non-overlapping setups~\cite{tribou2015multi} suffer from motion-dependent scale unobservability and drift in the absence of an IMU. On the Hilti--Trimble SLAM Challenge 2026~\cite{slamchallenge2026} rig, recent systems estimate the trajectory with feature-based visual-inertial SLAM~\cite{jeon2026acdc,jiang2026hilti}, while others recover the full trajectory offline through global factor-graph optimization~\cite{lee2026sqrtvins}. These systems, however, leave the trajectory in an arbitrary frame, neither localizing within a floor plan nor exploiting the one available on these construction sites to correct it.

\subsubsection*{\textbf{Localization with architectural priors}}
Several methods localize a camera within a pre-existing floor plan, aligning monocular detections or layout cues to it through semantic Monte-Carlo localization~\cite{mendez2018sedar} or learned ray-based filtering~\cite{chen2024f3loc}, returning a global position but without using the floor plan to improve the trajectory or the map. In the challenge's Localization task~\cite{slamchallenge2026}, Map-It Ralph~\cite{demirtas2026} likewise makes no use of the floor-plan geometry, reaching the plan frame from the provided anchor pose alone. The top three teams, in contrast, do use the floor plan, aligning a completed estimate to it offline: \textit{Z-FLoc} by a single global transform from a bird's-eye reconstruction~\cite{umemura2026z}, OmniRecon by 2D ICP on an offline structure-from-motion cloud~\cite{tanner2026}, and CUFE by refining completed trajectories for floor-plan consistency~\cite{cufe2026}. All of them, however, apply the floor plan only after the trajectory is complete, rather than integrating it into the estimation to correct drift as the trajectory is built.

\subsubsection*{\textbf{Architectural priors inside SLAM}}
Architectural plans can be leveraged inside SLAM systems to localize estimates within a building structure while offering an external metric reference to correct drift.
A recent system from the same challenge~\cite{zang2026} relies entirely on the benchmark's provided starting pose to localize within the building,
whereas real-world inspection tasks normally start from an estimated position, such as that obtained by having an operator match a few detected walls to the floor plan~\cite{bikandi2025bim}.
Furthermore, it uses the floor plan only to reject false loop-closure candidates, without directly correcting the drift.
Other works do integrate the floor plan into the estimation to correct drift, but sense the walls directly with a range or depth sensor.
LiDAR pose-graph methods anchor scans to a BIM through multi-session alignment~\cite{torres2023bim}, while~\cite{shaheer2023graph} matches an online scene graph of rooms and walls to one extracted from the floor plan, and \textit{ivS-Graphs}~\cite{bikandi2025bim} brings this to vision with a simpler, wall-only matching, from an RGB-D sensor. Such sensing is uncommon on hand-held site-capture rigs, and depth cameras reach only a few meters, short of a construction interior's spans. A depth-free system~\cite{driftFree} instead registers sparse map points to a digital twin, but its prior is a dense, photorealistic \textit{as-built} mesh that must be acquired separately and updated as the site evolves, unlike an \textit{as-planned} floor plan available from the design stage. Correcting drift online by integrating an \textit{as-planned} floor plan into a camera-only visual-inertial estimation remains unexplored.

\section{Methodology}
\label{sec:method}

\subsection{System Overview}
\label{sec:overview}
\system{} corrects the drift of a visual-inertial trajectory and localizes inside the building by introducing the \textit{as-planned} floor plan into a graph-based back-end. We build it upon ORB-SLAM3~\cite{campos2021orb}, a widely adopted and extensively benchmarked SLAM system of this kind.


\system{} takes four inputs: the two fisheye image streams, an IMU, a floor plan of the site, and one plan-frame initialization pose $\mathbf{T}_s$, needed only as a coarse estimate rather than ground truth, since later corrections absorb its error (Sec.~\ref{sec:backend}). It processes them through three modules, each supplying what the next requires (Fig.~\ref{fig:system_architecture}): a multi-camera visual-inertial SLAM (Sec.~\ref{sec:frontend-vslam}) estimates a continuous, metric trajectory and reconstructs a map; a wall detection and matching module (Sec.~\ref{sec:frontend-sem}) detects walls $W_s$ in the map and matches them to their floor-plan counterparts $W_{\mathrm{plan}}$;
and an integration module (Sec.~\ref{sec:backend}) folds each matched wall into a multi-stage back-end optimization that jointly re-estimates the trajectory and map, aligning them with the floor plan (Fig.~\ref{fig:front_image}).

\subsection{Multi-camera visual-inertial SLAM}
\label{sec:frontend-vslam}

\system{} extends monocular-inertial SLAM to the sensor setup of the rig. Facing opposite directions, the two fisheye cameras share too little overlap for stereo matching; each stream is tracked monocularly against a single shared map, with features from both hemispheres and the high-frequency IMU constraining one trajectory. The extension spans initialization, tracking, and map merging when the track is lost.

\textbf{Fast initialization.} A first metric map anchors the rest of the trajectory. Monocular-inertial pipelines build it only once enough parallax is available~\cite{vinsmono,campos2021orb}, which a sequence beginning at rest or turning in place does not provide, while stereo pipelines build it at once from a wide image overlap the rig lacks. In between, Li et al.~\cite{li2021robust} showed that, for multi-camera rigs with limited view overlap, the few features matched across cameras are enough to seed the map from a single frame. \system{} follows this idea, adapted to the rig and its IMU, trying two paths in order.
(a) Cross-camera metric seeding: Front-camera features falling in the narrow peripheral band shared by both cameras are searched in the back image along known cross-camera epipolar curves over a small set of depth hypotheses, and the median depth of the surviving matches sets the scale, every remaining feature placed at that depth along its ray. Unlike the vision-only method of~\cite{li2021robust} needing an accurately triangulated seed, this coarse map suffices: the $4$\,cm inter-camera baseline could not provide accurate depths, but the first inertial bundle adjustment recovers them together with the scale.

(b) Gyro-aided fallback: In visual odometry and outlier rejection, taking the inter-frame rotation from the gyroscope rather than estimating it has proven to make two-view geometry more robust~\cite{kneip2011robust,troiani20142}: it reduces the problem from five degrees of freedom to the two of the translation direction, recovered by a 2-point RANSAC~\cite{troiani20142}. \system{} applies this to initialization: when too few cross-camera matches survive, it falls back to two views with the rotation from gyro preintegration. This also removes the choice between a homography and a fundamental matrix, unreliable where planar surfaces dominate and parallax is small, which improves the robustness of the initialization in these challenging scenes.

\textbf{Multi-camera tracking and optimization.} The back camera is rigidly attached to the front one, so, as in other multi-camera visual-inertial systems~\cite{he2022towards}, its observations constrain the same visual-inertial state from the opposite viewing hemisphere. 
For a map point seen by the back camera, its reprojection residual composes the estimated front-camera pose with the fixed inter-camera extrinsic, known from the IMU-camera calibrations, before applying the Kannala--Brandt fisheye projection.
Each camera populates the map on its own, back-camera points being triangulated temporally between keyframes with the rig motion as baseline. A point is then searched only in the camera that created it while it stays in that field of view: as the two hemispheres barely overlap, searching it in both cameras would only add computational cost and cross-camera mismatches between similar structures. When it leaves, during turns or corridor reversals, it is projected into the opposite camera, which re-observes it and keeps it tracked across the hemisphere change. When both cameras end up mapping the same structure, fusion merges the duplicated points and culling removes the redundant ones.

\textbf{Inertial map merging.} Tracking can still be lost, on a fast turn or in front of a textureless wall~\cite{helmberger2022}. When the pose cannot be recovered in the current map, multi-map systems such as ORB-SLAM3 start a new one and connect it to the previous map only if place recognition later identifies an already mapped area~\cite{campos2021orb}. Because the floor plan is anchored to the first map through the initialization pose, a new map opened at its own origin stays outside the plan frame until the operator walks through an already mapped area again, which may never happen in a walkthrough without loops, and no plan correction can be applied to it meanwhile.

\system{} instead keeps the interrupted map and merges it with the new one through the IMU: the last keyframe before the loss is propagated by dead reckoning over the frames the loss lasts, and comparing this prediction with the first keyframe of the new map gives the transform between the two map frames. Both maps are metric and gravity-aligned, so the transform reduces to yaw and position, the four degrees of freedom unobservable in visual-inertial estimation~\cite{vinsmono}, and is applied once, leaving the internal geometry of each segment untouched. Both segments therefore remain in the anchored frame of the first map, so walls matched in either of them constrain the whole trajectory, and a structure seen on both sides of the loss refines the join itself (Sec.~\ref{sec:backend}).

\subsection{Semantic wall detection and floor-plan association}
\label{sec:frontend-sem}

This module supplies the back-end with wall correspondences in three steps: wall detection recovers the \textit{as-built} walls $W_s$, floor-plan processing extracts the plan walls $W_{\mathrm{plan}}$, and a drift-aware association matches the two.

\textbf{Wall detection.} Planar surfaces are established landmarks in structured-environment SLAM~\cite{kaess2015simultaneous}, but fitting them directly to a feature-based map is error-prone, since its points come from many structures and non-wall planar clutter such as cabinets or panels is easily mistaken for a wall. \system{} therefore detects walls semantically, classifying the map points and fitting wall planes only to those labelled as wall~\cite{bikandi2025bim}.
These labels come from a pretrained panoptic segmenter (EoMT~\cite{kerssies2025your}), which assigns every pixel of both fisheye images a COCO-panoptic class, a vocabulary that already contains the \textit{wall} class, so no site-specific fine-tuning is required.
Each local-map point is then assigned the semantic class of its projected pixel on a per-frame basis as observations arrive.
Labels are stabilized with a saturating hysteresis counter, so a point's label flips only under sustained contradictory evidence. Three filters then suppress spurious wall labels: a gravity filter demotes wall labels whose viewing normal is too close to vertical (floor/ceiling bleed-through), a depth band rejects unreliable projections, and newly created map points are withheld until their labels stabilize. 


At every keyframe, these wall-labelled points are fitted into wall planes with a sequential RANSAC in a gravity-aligned frame. Verticality removes one rotational degree of freedom, so a wall is parameterized by azimuth and distance only and fitted from a minimal sample of two points; hypotheses are scored by a robust RANSAC objective (MSAC) and accepted only if their inliers form a single contiguous wall segment.
A freshly fitted plane is tentative until confirmed by consistent re-fitting across several keyframes, after which it becomes a detected wall $W_s$ eligible for matching with the floor plan.

\textbf{Floor-plan processing.} Since the wall detection module recovers only the larger, clearly-observed walls, not every thin partition or facade detail drawn in the CAD, the floor plan is reduced to a comparable, canonical set of wall faces the detected walls can be matched against. \system{} converts the CAD wall layers into 2D wall faces, registers them to the plan raster by a grid search maximizing wall-pixel overlap, and canonicalizes them by merging collinear spans, removing duplicate or ambiguous faces, assigning consistent outward normals, and completing missing opposite faces. Exterior/interior labels are inherited from the CAD layers, with exterior taking precedence when a face combines both.

\textbf{Drift-aware association.} Each accepted match becomes a persistent constraint on the trajectory, so the association commits conservatively to avoid corrupting the rest of the estimate. The initialization pose $\mathbf{T}_s$ places the floor plan in the \textit{as-built} frame, so this step can find the correct associations between each detected wall $W_s$ and its floor-plan counterpart $W_{\mathrm{plan}}$.
A wall constrains the estimate only along its normal, so the integration stage (Sec.~\ref{sec:backend}) applies each correction anisotropically, and how much a candidate $W_{\mathrm{plan}}$ can be trusted then depends on how far the estimate has drifted along that direction. In principle, the estimator's pose covariance could measure this, but a keyframe-windowed bundle adjustment holds out-of-window poses fixed~\cite{orb_uncertainty}, so this covariance reflects only the local window and not this accumulated drift.

We therefore estimate this drift from the trajectory instead. It grows the longer the trajectory runs without a correction in that direction, so for each detected wall we track it through $\ell$, the length traveled since the last accepted exterior association with a parallel wall normal ($|\mathbf{n}_i^{\!\top}\mathbf{n}_j|\!\geq\!0.9$); only exterior wall matches, being the more reliable, reset $\ell$. A small $\ell$ means the perpendicular distance $\Delta d$ to a plan face is still reliable; a large $\ell$ means the trajectory may have drifted farther from the correct $W_{\mathrm{plan}}$. We use $\ell$ in two ways. First, it widens a hard distance gate: a candidate $W_{\mathrm{plan}}$ is admitted only if
\begin{equation}
\Delta d \;\leq\; \min\!\left(d_0 + \alpha_\ell\,\ell,\; c\right),
\label{eq:gate}
\end{equation}
so the tolerated perpendicular offset starts at $d_0$, grows with drift at rate $\alpha_\ell$, and is capped at a wall class dependent value $c$ that limits how far the gate can open (Table~\ref{tab:params}). 
This cap is tighter for interior walls than exterior ones ($c_{\mathrm{int}}\!<\!c_{\mathrm{ext}}$), because interior partitions are densely packed with near-parallel neighbours, where a looser gate risks matching the wrong twin, and are detected less reliably from shorter range, whereas well-separated facades can absorb more drift before a match becomes ambiguous. Second, it saturates the distance that enters the score, yielding the effective distance $\widetilde{\Delta d}$,
\begin{equation}
\widetilde{\Delta d} \;=\; (1-\sigma)\,\Delta d \;+\; \sigma\,\min(\Delta d,\, d_{\mathrm{sat}}),
\label{eq:sat}
\end{equation}
which equals the true distance $\Delta d$ when drift is small ($\sigma{=}0$) and caps it at $d_{\mathrm{sat}}$ when drift is large ($\sigma{=}1$). The blend factor $\sigma$ rises linearly from $0$ to $1$ as $\ell$ grows from $\ell_{\min}$ to $\ell_{\max}$ (Table~\ref{tab:params}), holding at $0$ below $\ell_{\min}$ and $1$ above $\ell_{\max}$. With large drift the distance therefore stops dominating, leaving the angle and overlap terms to decide among the remaining candidates.
After this drift-aware distance handling, each surviving plan face is scored with
\begin{equation}
s = w_a\,\Delta\theta + w_d\,\widetilde{\Delta d} + w_f\,(1-f),
\label{eq:score}
\end{equation}
where $s$ is the match cost, so the face minimizing $s$ wins; $\Delta\theta$ is the azimuth difference between the $W_s$ and $W_{\mathrm{plan}}$ normals, taken with sign so that a face whose normal points the opposite way is penalized rather than counted as aligned; $f\!\in\![0,1]$ is the fraction of the detected extent contained in the face; and the weights $w_a,w_d,w_f$ (Table~\ref{tab:params}) bring the radian, metre, and unitless terms to a common scale. The angle term is weighted most strongly because small orientation errors make a wall match unreliable even when its centroid is close.

The final match of a detected wall $W_s$ to its plan face $W_{\mathrm{plan}}$ is committed only if it is unambiguous, which a rejection cascade enforces before the pair is handed to the integration stage. The lowest-cost candidate is discarded if its score exceeds a class-dependent ceiling, if it does not beat the runner-up by a sufficient margin (Table~\ref{tab:params}), or if a same-class competitor lies at a comparable perpendicular distance.

\subsection{Multi-stage floor-plan integration}
\label{sec:backend}
\begin{figure}[t]
    \centering
    \includegraphics[width=0.9\columnwidth]{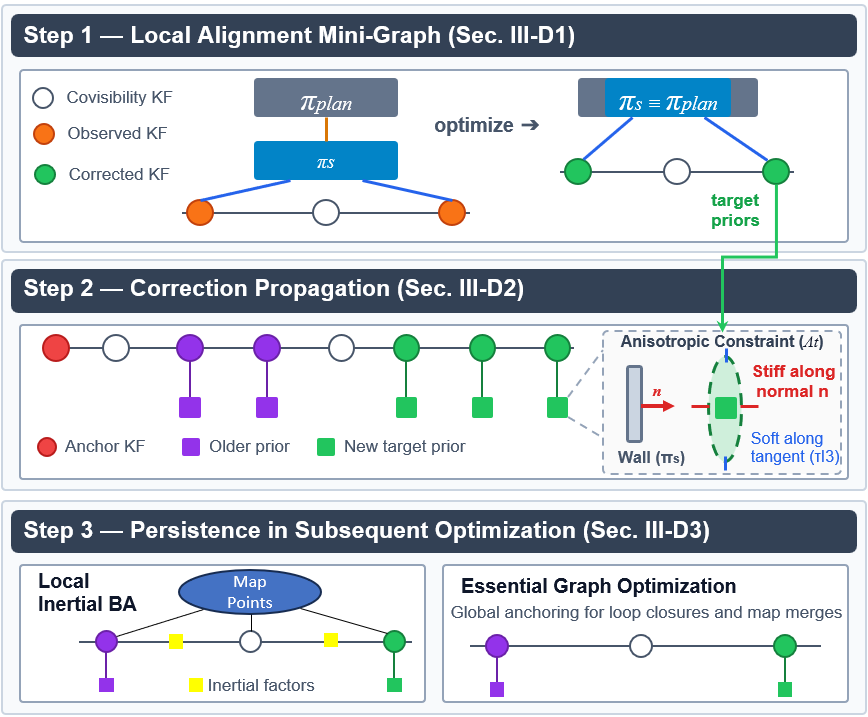}
    \caption{Incremental multi-stage integration of a matched wall, preserving earlier corrections.}
 \label{fig:steps}
\end{figure}

Each matched pair $(W_s,W_{\mathrm{plan}})$ provides a plan correspondence, but it reduces drift only once integrated into the SLAM graph as factors that optimize the trajectory and map. 
Introducing every accepted correspondence into one joint plan-constrained optimization makes that problem grow with the trajectory and accumulated associations~\cite{kaess2012isam2}. 
\system{} instead incorporates each correspondence incrementally through a multi-stage optimization run whenever a new association is committed. Because this re-estimation happens at every match, \system{} continually realigns and localizes within the floor plan, so errors such as an imperfect initial pose $\mathbf{T}_s$ are progressively absorbed. It runs in three steps: a local alignment over a bounded window (Step 1), a pose propagation of that correction (Step 2), and its persistence as a prior in later optimizations (Step 3), illustrated in Fig.~\ref{fig:steps}.

\subsubsection{Step 1: Local alignment mini-graph}
\label{sec:agraphba}

The first stage builds a local 4-DoF mini-graph for the newly matched pair $(W_s,W_{\mathrm{plan}})$, optimizing only yaw and translation because the wall constraints are vertical and roll/pitch are fixed by gravity. In it, the plan wall $W_{\mathrm{plan}}$ is a fixed vertex and the detected wall $W_s$ an optimizable one, joined by a high-information alignment edge; every keyframe observing $W_s$ is a pose vertex, tied to $W_s$ by a plane-projection edge and to its covisibility neighbours by relative 4-DoF edges. Aligning $W_s$ with $W_{\mathrm{plan}}$ therefore shifts the observing keyframes and removes their accumulated drift. Besides these observers, only the keyframes in the temporal gap between them and their covisibility neighbours are free to correct. If a wall was observed over a long span, this set is capped at a maximum window size (Table~\ref{tab:params}), keeping the most recent keyframes. 
Each match is corrected on its own window, keeping the update incremental and per-association.
While physical walls are finite surfaces represented by $W_s$ and $W_{\mathrm{plan}}$, their supporting infinite planes, $\pi_s$ and $\pi_{\mathrm{plan}}$, are used in this mini-graph formulation. The local alignment minimizes the joint cost:
\begin{equation}
\min_{\{\mathbf{T}_k\},\,\pi_s}\;
\norm{\pi_s\ominus\pi_{\mathrm{plan}}}^2_{\mathbf{\Lambda}_{B}}
+
\sum_{k\in\mathcal{O}}
\norm{\mathbf{e}^{\pi}_{k}}^2_{\mathbf{\Lambda}_{\pi}}
+
\sum_{(i,j)\in\mathcal{E}_{\mathrm{loc}}}
\norm{\mathbf{e}_{ij}}^2_{\mathbf{\Lambda}_{ij}},
\label{eq:local-cost}
\end{equation}
whose three terms weight these edges: the alignment edge ($\ominus$ the difference between infinite plane parameters~\cite{kaess2015simultaneous}) pulls the detected wall $\pi_s$ onto the fixed plan face $\pi_{\mathrm{plan}}$, penalizing their azimuth and signed-distance mismatch; plane-projection residuals $\mathbf{e}^{\pi}_{k}$ tie each observing keyframe $k\in\mathcal{O}$ to the wall measurement stored when it observed the wall, so consistency is enforced against the local observation, not a global drifted pose; and relative 4-DoF covisibility edges $\mathcal{E}_{\mathrm{loc}}$ preserve the local trajectory shape while yaw and translation adapt to the wall. The first keyframe is a gauge when no earlier plan correction exists.
Rather than modifying SLAM poses directly, the mini-graph outputs corrected keyframe targets storing their associated wall normals to guide subsequent directional priors

\subsubsection{Step 2: Correction propagation to the system}
\label{sec:propagation}


Applied on their own, the mini-graph targets would end at the window boundary, leaving a discontinuity against the rest of the trajectory. A second 4-DoF pose graph therefore propagates them to the surrounding keyframes. Its adaptive window is built around the newly corrected keyframes and an anchor, chosen when possible as the latest previous correction whose stored wall normal is parallel to the current one. The window bridges from this anchor to the newly corrected span and extends backward from it, farther if too few same-axis old priors fall inside, so the previous correction supports the update over multiple keyframes rather than a single fixed anchor. 
The graph combines relative edges (covisibility and temporal-chain) with unary priors (fresh targets from Step 1 and previous corrections) by minimizing the joint cost:
\begin{equation}
\min_{\{\mathbf{T}_k\}}
\sum_{(i,j)\in\mathcal{E}}
\norm{\mathbf{e}_{ij}}^2_{\mathbf{\Lambda}_{ij}}
+
\sum_{k\in\mathcal{P}}
\norm{\mathbf{e}_k}^2_{\mathbf{\Lambda}_k},
\label{eq:prop-cost}
\end{equation}
where $\mathcal{E}$ contains relative 4-DoF edges, $\mathcal{P}$ is the set of keyframes carrying targets from Step 1, $\mathbf{e}_{ij}$ is the relative residual, and $\mathbf{e}_k$ is the unary prior residual pulling keyframe $k$ toward its target pose, weighted by the block-diagonal information matrix $\mathbf{\Lambda}_k$ (Fig.~\ref{fig:steps}, Step~2 inset):
\begin{equation}
\begin{aligned}
\mathbf{\Lambda}_k &\;=\;
\begin{bmatrix}
\mathbf{\Lambda}_t & \mathbf{0}\\
\mathbf{0} & \lambda_r\,\mathbf{I}_3
\end{bmatrix}\in\mathbb{R}^{6\times6},\\[6pt]
\mathbf{\Lambda}_t &\;=\;
\tau\,\mathbf{I}_3
\;+\;
s\!\!\sum_{\mathbf{n}\in\mathcal{N}_k}
\mathbf{n}\mathbf{n}^{\!\top}\;\in\mathbb{R}^{3\times3},
\end{aligned}
\label{eq:aniso}
\end{equation}
The translation block $\mathbf{\Lambda}_t$ builds an anisotropic constraint based on environment geometry. For each observed wall normal $\mathbf{n} \in \mathcal{N}_k$, the outer-product term $s\,\mathbf{n}\mathbf{n}^{\!\top}$ injects high information $s$ perpendicular to the wall, while $\tau\mathbf{I}_3$ provides a small isotropic information floor in every direction. The prior therefore pins the keyframe stiffly across each wall it observed while leaving it free to slide along the wall surface and vertically. Keyframes matched to multiple non-parallel walls are naturally pinned along all their respective normals. The rotation block $\lambda_r\mathbf{I}_3$ provides an isotropic constraint on heading (yaw). Because roll and pitch remain fixed by gravity alignment in this 4-DoF formulation, yaw is the only rotational degree of freedom optimized, making a single scalar weight $\lambda_r$ sufficient.

\begin{table*}[t]
\vspace{0.3cm}
\setlength{\tabcolsep}{2pt}
\centering
\caption{Absolute Pose Error (APE) RMSE ($\mathrm{cm}$) on Hilti--Trimble SLAM Challenge 2026 sequences, as reported on the official challenge leaderboard on the day of the challenge. \textbf{Bold} and \underline{underline} denote best and second-best results per column per task; dashes (--) mark unavailable entries for Localization.}
\label{tab:ape}
\resizebox{\textwidth}{!}{%
\begin{tabular}{c|l|cc|cc|ccc|cc|cc|c|cccc|ccc|ccccc|c|c}
\toprule
\multicolumn{2}{c|}{} & \multicolumn{25}{c|}{\textbf{Floors and Sequences}} & \\
\cmidrule(lr){3-27}
\multicolumn{2}{c|}{} & \multicolumn{7}{c|}{\textbf{Ground Floors}} & \multicolumn{12}{c|}{\textbf{Upper Floors}} & \multicolumn{6}{c|}{\textbf{Underground Levels}} & \\
\cmidrule(lr){3-9} \cmidrule(lr){10-21} \cmidrule(lr){22-27}
\multicolumn{2}{c|}{}
 & \multicolumn{2}{c|}{\textbf{Floor 1}}
 & \multicolumn{2}{c|}{\textbf{Floor 2}}
 & \multicolumn{3}{c|}{\textbf{Floor EG}}
 & \multicolumn{2}{c|}{\textbf{Floor 3}}
 & \multicolumn{2}{c|}{\textbf{Floor 4}}
 & \multicolumn{1}{c|}{\textbf{Floor 5}}
 & \multicolumn{4}{c|}{\textbf{Floor 6}}
 & \multicolumn{3}{c|}{\textbf{Floor 7}}
 & \multicolumn{5}{c|}{\textbf{Floor UG1}}
 & \multicolumn{1}{c|}{\textbf{Floor UG2}}
 & \\
\midrule
\multicolumn{2}{c|}{}
 & \rotatebox{75}{07-07} & \rotatebox{75}{12-02}
 & \rotatebox{75}{12-02} & \rotatebox{75}{12-03}
 & \rotatebox{75}{10-16} & \rotatebox{75}{12-02a} & \rotatebox{75}{12-02b}
 & \rotatebox{75}{05-19} & \rotatebox{75}{12-02}
 & \rotatebox{75}{05-19} & \rotatebox{75}{12-02}
 & \rotatebox{75}{12-02}
 & \rotatebox{75}{06-18} & \rotatebox{75}{07-07} & \rotatebox{75}{12-02a} & \rotatebox{75}{12-02b}
 & \rotatebox{75}{12-02a} & \rotatebox{75}{12-02b} & \rotatebox{75}{12-03}
 & \rotatebox{75}{05-19} & \rotatebox{75}{06-18} & \rotatebox{75}{12-02a} & \rotatebox{75}{12-02b} & \rotatebox{75}{12-03}
 & \rotatebox{75}{12-02}
 & \rotatebox{75}{\textbf{Average}} \\
\midrule
\multicolumn{2}{r|}{\textbf{Duration ($\mathrm{sec.}$)}} & 133.7 & 276.3 & 143.7 & 149.9 & 242.9 & 126.4 & 162.5 & 115.5 & 132.7 & 91.9 & 193.4 & 160.5 & 69.3 & 73.0 & 170.7 & 124.6 & 117.9 & 155.1 & 198.3 & 205.5 & 165.8 & 254.8 & 219.7 & 134.4 & 223.1 & 161.7 \\
\multicolumn{2}{r|}{\textbf{Length ($\mathrm{m}$)}} & 157.8 & 321.8 & 154.8 & 138.8 & 240.4 & 114.9 & 158.6 & 128.2 & 148.8 & 97.8 & 214.1 & 174.9 & 79.4 & 72.2 & 210.6 & 145.8 & 145.4 & 197.1 & 152.6 & 262.5 & 222.6 & 351.7 & 322.5 & 119.1 & 292.1 & 185.0 \\
\midrule
\multirow{6}{*}{\rotatebox[origin=c]{90}{\textbf{Localization}}} & Map-It Ralph~\cite{demirtas2026} & 71.92 & 39.68 & 577.06 & 233.11 & 41.86 & 87.98 & 320.64 & 103.89 & 182.71 & 38.67 & 49.77 & 82.38 & 49.74 & 41.31 & 50.59 & 109.38 & 301.10 & 105.34 & 69.70 & 95.13 & 147.77 & 264.79 & 88.25 & 127.67 & -- & 136.69 \\
 & CUFE~\cite{cufe2026} & 66.98 & 41.20 & 60.07 & 42.64 & 25.54 & 30.83 & \textbf{18.22} & 84.93 & 106.07 & 75.80 & 45.97 & 65.83 & \textbf{20.80} & \underline{28.94} & 41.87 & 78.11 & 63.01 & 88.80 & 76.51 & 38.88 & 66.50 & 48.09 & 83.49 & 142.69 & -- & 60.07 \\
 & OmniRecon~\cite{tanner2026} & 38.05 & 30.51 & 38.79 & 45.59 & 33.32 & 19.88 & 41.99 & 40.12 & 64.74 & 44.20 & \underline{17.58} & 53.57 & 30.20 & 40.37 & 52.20 & 27.42 & \underline{14.26} & 74.43 & 27.60 & 46.85 & \underline{55.00} & 58.41 & \underline{53.50} & 58.08 & -- & 41.94 \\
 & \textit{Z-FLoc}~\cite{umemura2026z} & \underline{28.87} & \textbf{17.78} & \textbf{32.55} & \underline{29.98} & \textbf{15.07} & \textbf{12.48} & \underline{18.37} & \underline{17.54} & \underline{36.45} & \underline{24.22} & \textbf{12.60} & \textbf{14.31} & 27.69 & 33.57 & \textbf{16.10} & \underline{19.08} & \textbf{13.30} & \textbf{21.04} & \underline{22.61} & \underline{32.17} & \textbf{33.56} & \textbf{24.25} & \textbf{42.62} & \textbf{23.77} & -- & \textbf{23.75} \\
 & \textbf{Ours (\system{})} & \textbf{17.69} & \underline{25.48} & \underline{33.57} & \textbf{18.54} & \underline{22.10} & \underline{13.48} & 37.79 & \textbf{13.65} & \textbf{31.77} & \textbf{17.37} & 21.30 & \underline{31.65} & \underline{21.56} & \textbf{18.85} & \underline{34.09} & \textbf{18.58} & 26.07 & \underline{31.57} & \textbf{21.30} & \textbf{27.35} & 79.10 & \underline{46.90} & 62.10 & \underline{31.19} & -- & \underline{29.29} \\
 & Ours (w/o plan) &131.48 & 251.39 & 151.83 & 95.04 & 243.66 & 26.28 & 100.84 & 55.17 & 50.33 & 100.81 & 244.43 & 54.88 & 58.60 & 51.21 & 75.51 & 35.91 & 163.64 & 113.40 & 98.81 & 220.39 & 206.57 & 350.86 & 483.25 & 109.04 & -- & 144.72 \\
\midrule
\multirow{6}{*}{\rotatebox[origin=c]{90}{\textbf{SLAM}}} & ACDC-VSLAM~\cite{jeon2026acdc} & \textbf{10.07} & \textbf{9.82} & \textbf{7.69} & \textbf{11.03} & \textbf{7.84} & \textbf{5.53} & \underline{6.70} & 12.60 & \textbf{6.50} & 8.01 & \underline{7.12} & \underline{5.47} & 6.36 & \textbf{5.82} & \underline{6.02} & \textbf{5.37} & \textbf{5.95} & \textbf{5.82} & \textbf{10.27} & \underline{15.36} & \underline{19.88} & \textbf{11.99} & \textbf{13.17} & \underline{6.24} & \textbf{12.86} & \textbf{8.94} \\
 & $\sqrt{\text{VINS}}$~\cite{lee2026sqrtvins} & \underline{14.65} & \underline{10.60} & \underline{13.54} & 18.83 & 9.87 & \underline{7.70} & \textbf{6.67} & \textbf{9.07} & \underline{8.66} & \underline{7.83} & 9.18 & 8.37 & \underline{5.88} & \underline{6.93} & 8.83 & 8.82 & \underline{8.89} & 7.49 & \underline{11.34} & \textbf{13.18} & \textbf{18.97} & \underline{16.16} & \underline{16.43} & 10.93 & \underline{13.67} & \underline{10.90} \\
 & Undisclosed & 17.59 & 11.76 & 36.04 & \underline{12.55} & \underline{9.46} & 8.99 & 12.75 & 12.86 & 36.01 & 12.97 & \textbf{6.74} & \textbf{5.43} & 16.71 & 14.72 & \textbf{5.41} & \underline{7.25} & 8.92 & \underline{7.46} & 16.32 & 27.20 & 33.24 & 49.54 & 50.03 & \textbf{6.13} & 40.83 & 18.68 \\
 & QQ~\cite{jiang2026hilti} & 29.42 & 28.59 & 23.23 & 15.14 & 10.97 & 20.88 & 20.48 & 12.29 & 38.15 & \textbf{6.78} & 16.40 & 12.61 & \textbf{5.32} & 9.14 & 12.36 & 9.03 & 13.18 & 19.41 & 23.93 & 24.08 & 39.23 & 28.89 & 39.67 & 16.66 & 28.34 & 20.17 \\
 & \textbf{Ours (\system{})} & 18.35 & 17.93 & 19.19 & 15.88 & 18.15 & 8.93 & 20.20 & \underline{11.95} & 25.05 & 13.92 & 15.72 & 33.58 & 17.05 & 10.32 & 29.84 & 10.50 & 18.31 & 11.49 & 12.35 & 30.98 & 56.87 & 56.47 & 83.06 & 20.52 & 29.17 & 24.23 \\
 & Ours (w/o plan) &104.96 & 128.91 & 50.79 & 78.35 & 116.47 & 10.40 & 22.32 & 35.78 & 93.05 & 55.04 & 100.69 & 21.00 & 29.62 & 14.69 & 49.29 & 18.78 & 37.41 & 46.37 & 32.66 & 155.01 & 68.01 & 177.93 & 187.25 & 74.11 & 121.42 & 73.21 \\
\bottomrule
\end{tabular}%
}
\begin{minipage}{\textwidth}
\vspace{3pt}
\footnotesize
Full challenge leaderboard: \url{https://hilti-trimble-challenge.com/leaderboard-2026}
\end{minipage}
\end{table*}

After propagation, the optimized poses are written back to the SLAM state. Map points move rigidly with their most recent corrected observer, velocities rotate with their keyframes, and the IMU preintegration terms are re-evaluated at the new states. Without this patch, the next inertial local BA would see inconsistent inertial residuals and pull the trajectory back toward the pre-correction state.

\subsubsection{Step 3: Persistence in subsequent optimization}
\label{sec:persistence}

A correction is useful only if it persists in the optimizations that follow. We thus store each corrected target with its wall-normal set $\mathcal{N}_k$ and reintroduce it as a unary anisotropic prior whenever the affected keyframe appears in a later optimization.

In the local inertial BA that follows each correction, the priors keep the continuous SLAM estimate coherent with them, using the same anisotropic form but a softer information ratio (Table~\ref{tab:params}) so that reprojection and inertial terms still refine the map along the wall while the floor plan prevents drift back across its normal. 
Each local BA jointly optimizes map points, keyframe poses, velocities, and IMU biases under these plan priors, so the final map and trajectory jointly satisfy visual, inertial, and structural constraints.

To enforce global alignment without noise, floor-plan priors enter the essential-graph optimization~\cite{campos2021orb} at loop closures and map merges. These constraints are applied only to keyframes associated with exterior walls, more reliable than interior ones.

\section{Experimental Results}
\label{sec:eval}

\begin{table}[b]
\caption{Key parameters of \system{}.}
\label{tab:params}
\centering
\footnotesize
\setlength{\tabcolsep}{2pt}
\begin{tabular}{@{}llll@{}}
\toprule
Parameter & Symbol & Value & Description \\
\midrule
\multicolumn{4}{@{}l}{\textit{Drift-aware association (Sec.~\ref{sec:frontend-sem})}} \\
score weights & $w_{a,d,f}$ & $6$\,rad$^{-1}$, $1$\,m$^{-1}$, $1$ & angle, dist., overlap \\
gate ramp & $d_0,\alpha_\ell$ & $1$\,m, $0.1$ & offset, drift growth \\
gate cap & $c_{\mathrm{ext/int}}$ & $5$, $2.5$\,m & ext./int. distance cap \\
dist. saturation & $d_{\mathrm{sat}}$ & $1$\,m & score distance cap \\
saturation range & $\ell_{\min/\max}$ & $15$, $40$\,m & score ramp \\
runner-up margin & -- & $1.6\times$ & match uniqueness \\
\midrule
\multicolumn{4}{@{}l}{\textit{Floor-plan integration (Sec.~\ref{sec:backend})}} \\
mini-graph window & -- & $200$\,KF & max.\ KFs in Step1 \\
along-normal info & $s$ & $10^4$ & stiffness across wall \\
tangent floor & $\tau$ & $1$ / $50$ & propagation / local BA \\
rotation info & $\lambda_r$ & $10^3$ / $10^4$ & propagation / local BA \\
\bottomrule
\end{tabular}
\end{table}

\subsection{Validation Methodology}
\label{sec:eval-setup}

\textbf{Datasets.} We evaluate on the Hilti--Trimble SLAM Challenge 2026 dataset~\cite{slamchallenge2026}, recorded in active construction sites with a hand-held rig carrying two back-to-back $\sim$200$^\circ$ fisheye cameras operating at 30\,Hz and a 1\,kHz IMU.
We address both official tasks with the same system. Localization is scored in the floor-plan frame and SLAM is scored after rigid alignment to the reference. Localization contains 24 scored sequences, excluding Floor~UG2 because no floor plan is released.


We group the sequences into ground floors, upper floors, and underground levels. Across all of them, the site is under construction and can differ from the floor-plan. The ground floors range from open spaces to the more built-out entrance level EG; the upper floors include interior areas and one outdoor terrace run; and the underground levels are large, sparsely-walled spaces dominated by structural columns.

\textbf{Baselines.} We compare \system{} against the other top-five competitors of each task (Table~\ref{tab:ape}), with per-sequence results taken from the official challenge release~\cite{slamchallenge2026}.

\textbf{Implementation Details.} \system{} runs on a workstation equipped with an Intel Core i9-11950H CPU and an NVIDIA T600 GPU ($4$\,GB), 32\,GB RAM.
Key parameters are listed in Table~\ref{tab:params}.

\begin{figure*}[!t]
  \centering
  \includegraphics[width=0.95\textwidth]{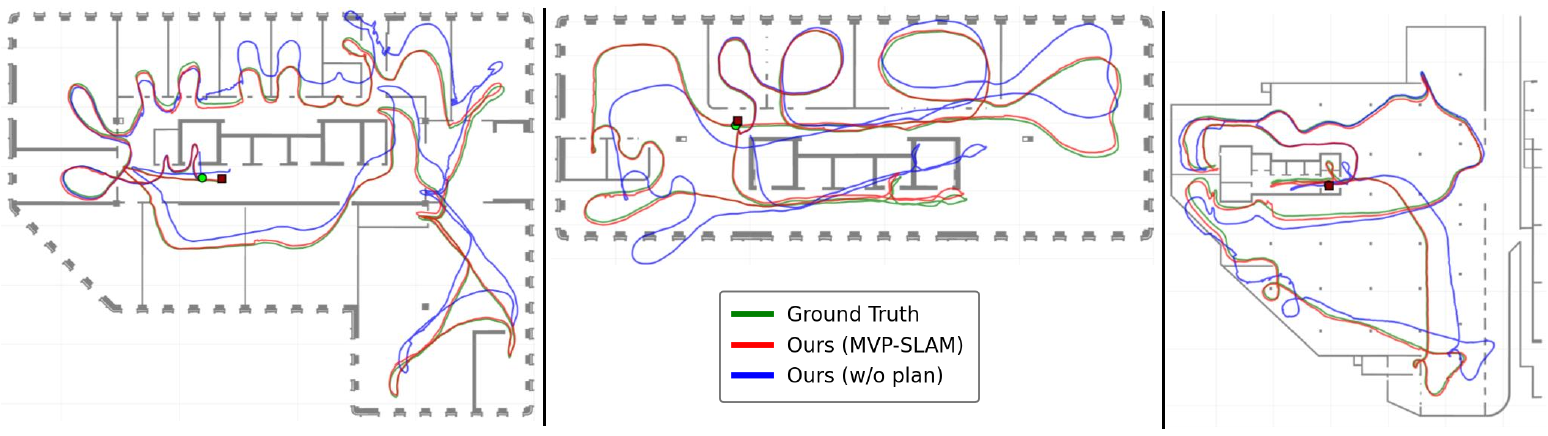}

  \vspace{-2.2em}

  \begin{subfigure}[t]{0.33\textwidth}\caption{Floor EG (10-16)}\label{fig:traj-eg}\end{subfigure}\hfill
  \begin{subfigure}[t]{0.33\textwidth}\caption{Floor 4 (12-02)}\label{fig:traj-f4}\end{subfigure}\hfill
  \begin{subfigure}[t]{0.31\textwidth}\caption{Floor UG1 (12-02a)}\label{fig:traj-ug1}\end{subfigure}

  \vspace{-0.25em} 
  \caption{Trajectory comparison in the \textit{Localization} task, in the floor-plan frame, on three representative sequences. 
  }
  \label{fig:trajectories}
\end{figure*}

\textbf{Trajectory estimation Performance.} The challenge provides a LiDAR-inertial reference trajectory, used only for evaluation. Both tasks report per-sequence RMSE of the Absolute Pose Error (APE) against it: 2D in the floor-plan plane for Localization, 3D after rigid alignment for SLAM. A trajectory must cover at least $99\%$ of the reference poses to be scored.
To isolate the impact of map priors, we also evaluate an ablated variant, \textit{Ours (w/o plan)}. It uses the same visual-inertial SLAM setup but disables wall associations and floor-plan integration.

\textbf{SLAM initialization and robustness ablation.} We assess the fast metric initialization and the map preservation across tracking losses of the multi-camera front-end (Sec.~\ref{sec:frontend-vslam}) with an ablation. The reduced configuration disables them, so it initializes and recovers as a standard visual-inertial SLAM would, while the full configuration keeps them enabled.
We measure initialization success, time to metric scale, trajectory coverage, and whether the run satisfies the challenge coverage requirement on six sequences, two per environment type.




\textbf{Wall detection and matching Performance.} We evaluate wall detection and wall-to-plan matching on the same six sequences, using precision, recall, and F1. The reference set is hand-labeled from the floor plan, camera images, and reconstructed map, and contains only walls that are visible to the cameras and supported by reconstructed map points. Unbuilt planned walls and transient clutter are ignored.
For detection, a true positive is a detected plane on the same physical wall as a reference wall; a false positive is a plane fitted to non-wall structure, clutter, or empty space; and a false negative a reference wall not recovered. For matching, we manually label each detected wall's correct plan face; a match is correct only when the committed association links to it.

\subsection{Results and Discussion}
\label{sec:eval-results}


\subsubsection{\textbf{Trajectory Estimation Performance}}
Table~\ref{tab:ape} reports the APE RMSE for both official challenge tasks. In the \textit{Localization} task, \system{} achieves a mean APE of $0.29$\,m and ranks second among the 22 Localization teams, close to the winning \textit{Z-FLoc} submission ($0.24$\,m). \system{} is best or second-best on most sequences, showing that the plan constraints keep the trajectory well aligned with the building frame across floors and environment types.
Without plan integration, \textit{Ours (w/o plan)} reaches $1.45$\,m mean APE; while tracking remains stable, accumulated drift is no longer bounded.
Adding the plan integration reduces the Localization error by $4.9\times$, from $1.45$\,m to $0.29$\,m. Fig.~\ref{fig:trajectories} illustrates this effect on representative ground-floor, upper-floor, and underground sequences.

In the \textit{SLAM} task, \system{} obtains $0.24$\,m mean APE and ranks fifth among 62 teams. As this metric only accounts for the trajectory shape (Sec.~\ref{sec:eval-setup}), the top-ranked systems, led by ACDC-VSLAM~\cite{jeon2026acdc} ($0.09$\,m) and $\sqrt{\text{VINS}}$~\cite{lee2026sqrtvins} ($0.11$\,m), reach lower APE through mechanisms such as enhanced loop closure, effective since the challenge sequences often revisit previously seen areas, and richer visual features. \system{} instead uses the floor plan to both localize the trajectory and correct drift, reaching the top five and demonstrating that the floor-plan integration reduces the aligned error $3.0\times$ from $0.73$\,m (\mbox{Ours w/o plan}) to $0.24$\,m.
As established in Sec.~\ref{sec:sota}, the other Localization teams apply the floor plan offline, once the trajectory is built~\cite{umemura2026z,tanner2026,cufe2026}; \system{} is instead the only one to integrate it online, folding each match into the estimation as it is detected. The same distinction holds in the \textit{SLAM} task, where \system{} is again the top-ranked team among those that integrate the floor plan online, and the only one to also localize within it, showing the building's floor plan is a promising, distinct prior for improving a SLAM system.

\begin{table}[b]
\caption{Initialization and robustness ablation for \mbox{Ours} \mbox{(w/o plan)}: 3 repeats per sequence, both configurations initialized every run. $t_{\mathrm{metric}}$: median time to metric scale (s); Cov.: mean coverage (\%); APE: median Localization error (m).}
\label{tab:frontend}
\centering
\footnotesize
\setlength{\tabcolsep}{3pt}
\begin{tabular}{@{}l l ccc ccc@{}}
\toprule
& & \multicolumn{3}{c}{\textbf{Reduced}} & \multicolumn{3}{c}{\textbf{Full}} \\
\cmidrule(lr){3-5}\cmidrule(lr){6-8}
\textbf{Floor} & Sequence & $t_{\mathrm{metric}}$ & Cov.\ & APE & $t_{\mathrm{metric}}$ & Cov.\ & APE \\
\midrule
\multirow{2}{*}{\textit{Ground}}
& Floor 1 (07-07)    & $3.1$ & $100$  & $1.38$ & $2.5$ & $100$ & $1.31$ \\
& Floor EG (10-16)   & $2.5$ & $100$  & $3.02$ & $2.2$ & $100$ & $2.44$ \\
\midrule
\multirow{2}{*}{\textit{Upper}}
& Floor 3 (05-19)    & $2.8$ & $100$  & $0.58$ & $2.5$ & $100$ & $0.55$ \\
& Floor 6 (12-02a)   & $2.5$ & $100$  & $1.01$ & $2.5$ & $100$ & $0.76$ \\
\midrule
\multirow{2}{*}{\textit{Under}}
& Floor UG1 (06-18)  & $6.6$ & $96.7$ & $28.0$ & $2.5$ & $100$ & $2.07$ \\
& Floor UG1 (12-02a) & $8.1$ & $100$  & $5.10$ & $2.4$ & $100$ & $3.51$ \\
\bottomrule
\end{tabular}
\end{table}

\subsubsection{\textbf{SLAM Initialization and Robustness Ablation}}
Table~\ref{tab:frontend} shows that both configurations initialize reliably and reach metric scale in $\sim$$2.5$s on upper floors, and diverge only in underground environments. Here, the reduced configuration takes up to $3\times$ longer to reach metric scale ($6.6$–$8.1$s) and fails the $99\%$ coverage threshold on Floor~UG1 (06-18).  This degradation occurs when a hard initialization briefly drops tracking, creating a second map frame. Without map preservation, the two maps remain unmerged. As a result, abandoned keyframes drop out of the trajectory, causing task failures and misaligning the floor plan reference. By maintaining continuity across tracking losses, map preservation merges these segments, recovering full coverage and reducing Floor~UG1 (06-18) Localization error from $28$m to $2.07$m.

\begin{table}[t]
\vspace{0.25cm}
\caption{Wall-detection and wall-matching performance (precision P, recall R, F1, in percent). Overall pools counts.}
\label{tab:wall}
\centering
\footnotesize
\setlength{\tabcolsep}{3pt}
\begin{tabular}{@{}l l cccccc@{}}
\toprule
& & \multicolumn{3}{c}{\textbf{Detection}} & \multicolumn{3}{c}{\textbf{Matching}} \\
\cmidrule(lr){3-5}\cmidrule(lr){6-8}
\textbf{Floor} & Sequence & P & R & F1 & P & R & F1 \\
\midrule
\multirow{2}{*}{\textit{Ground}} 
& Floor 1 (07-07)    & 94.4  & 50.0 & 65.4 & 100.0 & 94.1  & 97.0 \\
& Floor EG (10-16)   & 100.0 & 28.6 & 44.4 & 100.0 & 100.0 & 100.0 \\
\midrule
\multirow{2}{*}{\textit{Upper}} 
& Floor 3 (05-19)    & 100.0 & 57.9 & 73.3 & 100.0 & 100.0 & 100.0 \\
& Floor 6 (12-02a)   & 84.6  & 35.5 & 50.0 & 90.9  & 100.0 & 95.2  \\
\midrule
\multirow{2}{*}{\textit{Under}} 
& Floor UG1 (06-18)  & 100.0 & 37.8 & 54.9 & 100.0 & 100.0 & 100.0 \\
& Floor UG1 (12-02a) & 100.0 & 40.4 & 57.6 & 100.0 & 100.0 & 100.0 \\
\midrule
& \textbf{Overall}   & \textbf{96.9} & \textbf{38.4} & \textbf{55.0} & \textbf{98.9} & \textbf{98.9} & \textbf{98.9} \\
\bottomrule
\end{tabular}
\end{table}

\subsubsection{\textbf{Wall Detection and Matching Performance}}
Table~\ref{tab:wall} reports wall detection and wall-to-plan matching on six representative sequences. Detection is conservative by design: it reaches high precision ($96.9\%$) but moderate recall ($38.4\%$). This choice follows from the integration design (Sec.~\ref{sec:backend}), where each accepted wall becomes a persistent plan prior; a wrong wall is therefore more harmful than a missed one. Most missed walls are visible in the images but have too few or too noisy reconstructed map points to support a stable plane fit, especially in cluttered or texture-scarce areas.

Once a wall is detected, association is highly reliable: matching precision, recall, and F1 are all $98.9\%$ pooled. The detected wall candidates are usually close to their true plan faces and separated from alternatives, so the score (Eq.~\ref{eq:score}) and rejection cascade resolve most matches cleanly. The single wrong match occurs on Floor~6, where two nearly collinear plan faces fall within the gate, and the single missed match is a detected wall that receives no association.
Thus, in these sequences, the plan prior is limited mainly by which walls can be detected from the sparse map, not by the association step. This is sufficient for localization when the accepted walls are distributed: Floor~EG, for example, attains a $22$\,cm Localization APE despite only $28.6\%$ wall-detection recall.


\textbf{Limitations.} \system{} corrects drift only where it detects a wall and matches it to the floor plan. Its wall module fits planar walls, so in open areas with few walls, notably large underground spaces dominated by structural columns, few matches are available and the trajectory drifts uncorrected.

\section{Conclusions and Future Work}
\label{sec:conclusions}

We presented \system{}, a multi-camera visual-inertial SLAM system that uses an \textit{as-planned} floor plan as a reference to reduce drift accumulated on construction sites. It tracks its position on two opposite-facing fisheye cameras and an IMU, and corrects the drift online by mapping walls, matching them to the floor plan, and integrating it into the SLAM back-end through each match as the trajectory is built. On the Hilti--Trimble SLAM Challenge 2026 it ranked \textbf{second} in Localization ($0.29$\,m mean RMSE) and \textbf{fifth} in SLAM ($0.24$\,m). It is the top team in both tasks among those that also operate online, integrate the floor plan, and localize within it.

Future work will explore lightweight map densification to make the detection of walls and additional structural primitives such as columns easier, integrating them into the optimization for drift correction. It will also target the SLAM's internal consistency directly, strengthening loop closure and adopting the richer-feature mechanisms of higher-ranked SLAM teams.

\bibliographystyle{IEEEtran}
\bibliography{root}

\end{document}